323# Cautious Propagation in Bayesian Networks *

**Finn V. Jensen**
Aalborg University
Dept. of Math. & Computer Science
DK-9000 Aalborg
E-mail: fvj@iesd.auc.dk## Abstract

Consider the situation where some evidence $e$ has been entered to a Bayesian network. When performing conflict analysis, sensitivity analysis, or when answering questions like "What if the finding on $X$ had been $y$ instead of $x$?", you need probabilities $P(e' \mid h)$ where $e'$ is a subset of $e$, and $h$ is a configuration of a (possibly empty) set of variables.

Cautious propagation is a modification of HUGIN propagation into a Shafer-Shenoy-like architecture. It is less efficient than HUGIN propagation; however, it provides easy access to $P(e' \mid h)$ for a great deal of relevant subsets $e'$.

Keywords: Bayesian networks, propagation, fast retraction, sensitivity analysis.## 1 Introduction

As an example for motivating the introduction of yet another propagation method, consider the junction tree in Figure 1, with evidence $e = \{s, t, u, v, w, x, y, z\}$ entered as indicated.

Suppose you want to perform a conflict analysis (Jensen, Chamberlain, Nordahl & Jensen 1991). Then you first calculate

$$\mathrm{conf}(e) = \log \frac{P(r)P(s)P(t)P(u)P(v)P(w)P(x)P(y)P(z)}{P(e)}$$

If conf is well above 0, it is an indicaiton that the evidence is conflicting. The next thing you would like to do is to trace the conflict further down. You may for example want to "retract" some of the evidence and calculate the conflict measure for the remaining evidence. You may also want to investigate whether it is the group of findings $\{x, y, z\}$ which is in conflict with the rest. To do so you need probabilities of the type $P(e')$, where $e'$ is a subset of $e$. There is a straightforward way to calculate these probabilities, namely to enter $e'$ as evidence, and then most propagation algorithms will provide $P(e')$ as a side effect. However, this is clumsy and time-consuming. Therefore you may be willing to pay an overhead in time and space for a propagation method which gives you a more direct access to probabilities of subsets.

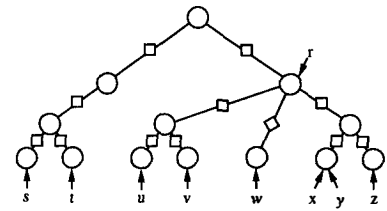

Figure 1: A junction tree with evidence $e = \{r, s, t, u, v, w, x, y, z\}$ entered. The circles are nodes consisting of sets of variables, and the boxes are separators consisting of the intersection of the neighbouring nodes. The top node is the chosen root.

In this paper we shall present a propagation method which we call *cautious propagation*. It provides very easy access to $P(e')$ for more subsets than any other propagation method known to the author, and extended with *cautious evidence entering* it provides a very fast way of calculating the impact on a hypothesis if a finding as retracted or changed from one value to another.

Cautious propagation is a modification of HUGIN propagation (Jensen, Lauritzen & Olesen 1990) to a message passing design like Shafer-Shenoy propagation (Shafer & Shenoy 1990). We call it cautious propagation because it works in a way not destroying the original tables in the junction tree. Cautious evidence entering was proposed by Dawid (1992) as *fast retrac-* *tion*, however, in his setting it only provided fast retraction in special cases.

We start with a closer analysis of the HUGIN propagation algorithm to pin-point its drawback with respect to calculation of $P(e')$s, and cautious propagation is defined together with its proof of correctness. Then cautious evidence entering is defined, and we compare the time and space complexity of HUGIN and cautious propagation. Finally, it is illustrated how cautious propagation can be used for sensitivity analysis.

## 2  $P(e')$s provided by HUGIN propagation

HUGIN propagation is performed in two phases, a *CollectEvidence* phase, where messages are sent from the leaves to a chosen root, and a *DistributeEvidence* phase, where messages are reversed. The content of the messages is explained below.

Consider the situation in Figure 2. Assume that the junction tree is consistent before the evidence is entered (for each clique $V$ its table is $P(V)$, and for each separator $S$ its table is $P(S)$), and let the evidence $e$ be divided by $S$ into $e_l$ and $e_r$.

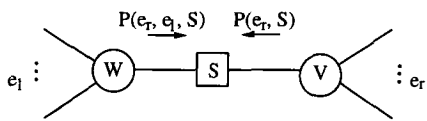

Figure 2: The transmitted tables in a network where *CollectEvidence* is called at the left of $W$.

If *CollectEvidence* is called in a node at the left of $W$, the table to be transmitted from $V$ to $S$ is $P(e_r, S)$, and $\frac{P(e_r, S)}{P(S)}$ is transmitted further to $W$ to be multiplied to $P(W)$. When later *DistributeEvidence* is activated, the table transmitted from $W$ to $S$ is $P(e_r, e_l, S)$ and $\frac{P(e_r, e_l, S)}{P(e_r, S)}$ is transmitted to $V$ and multiplied to $V$'s table ($\frac{0}{0}$ is set to 0).

This gives us a way to calculate probabilities for subsets of the evidence at each separator. Since $S$ initially holds $P(S)$, we have at our disposal at $S$ the three tables $P(S), P(e_r, S)$ and $P(e_r, e_l, S)$. Then

$$P(e_r) = \sum_S P(e_r, S)$$

Since $e_r$ is independent of $e_l$ given $S$ we have

$$P(e_r \mid S)P(e_l \mid S) = P(e_r, e_l \mid S) = \frac{P(e_r, e_l, S)}{P(S)}$$

This yields a formula for the calculation of $P(e_l)$:

$$P(e_l) = \sum_S \frac{P(e_r, e_l, S)P(S)}{P(e_r, S)}$$

Unfortunately it can only be used if $P(e_r, s) \neq 0$ for all states $s$ of $S$. If $P(e_r, s) = 0$ the fraction is put at 0, and the sum is only a lower bound for $P(e_l)$ (If we for these states put the fraction to $P(s)$ we also get an upper bound for $P(e_l)$). The problem is that the clique tables $P(V)$ are affected by the propagation, and if 0's are introduced you cannot restore the table.

## 3  Cautious propagation

Consider Figure 2 again and observe that in the *DistributeEvidence* phase, $P(V, e_r)$ is multiplied by $\frac{P(e_r, e_l, S)}{P(e_r, S)}$. If instead we multiply by $P(e_l \mid S)$ the only difference will be multiplications for configurations $s$, where $P(s, e_r) = 0$. In that case the corresponding entries in $P(V, e_r)$ are also 0. So multiplying by $P(e_l \mid S)$ will give exactly the same result as the HUGIN propagation will give.

Therefore, the idea behind cautious propagation is that in the *CollectEvidence* phase let $S$ keep $P(S)$, and store $P(W)$ and $P(e_r \mid S)$ separately near $W$. Now, assume that $W$ has received and stored the messages $P(e_1 \mid S_1), \ldots, P(e_n \mid S_n)$ (see Figure 3) where $e_w = e_1 \cup \ldots \cup e_n$.

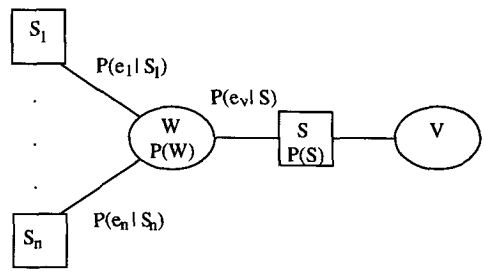

Figure 3: The situation when $W$ is about to pass $P(e_w, S)$ to $S$.

Then

$$P(e_w, S) = \sum_{W \setminus S} P(W) P(e_1 \mid S_1) \ldots P(e_n \mid S_n)$$

and

$$P(e_w \mid S) = \frac{P(e_w, S)}{P(S)}$$

If $W$ is the root, Figure 3 describes the situation after *CollectEvidence*($W$), and recursively it will in the *DistributeEvidence* phase be the situation for all non-leaf nodes.

A more precise description of cautious propagation is the following:

Before evidence is entered, each separator $S$ holds $P(S)$ and each clique $V$ holds $P(V)$. Whenever $S$ receives a table $T_V(S)$ from a neighbour $W$, it is divided by $P(S)$, stored, and a message is sent to the other neighbour clique $V$ (see Figure 4).

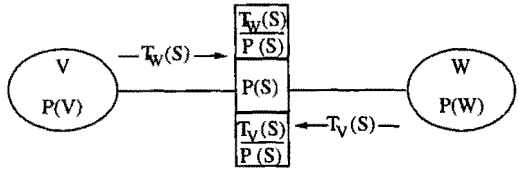

Figure 4: A link in a junction tree and the tables stored when a message has been passed in both directions.

A clique $W$ can send a table to a neighbour separator $S$ if it has received a message from all its other neighbouring separators $S_1, \ldots, S_n$. The table sent is

$$T_V(S) = \sum_{W \setminus S} P(W) \frac{T_W(S_1)}{P(S_1)} \cdots \frac{T_W(S_n)}{P(S_n)}$$

For the correctness of the method we see from the considerations above that it is enough to show that $T_V(S) = P(S, e_W)$, where $e_W$ is the evidence entered at the subtree containing $S$ and $W$, but not $V$:

If $W$ is a leaf then

$$T_V(S) = \sum_{W \setminus S} P(W, e_W) = P(S, e_W)$$

If $T_W(S_i) = P(S_i, e_i)$ then $\frac{T_W(S_i)}{P(S_i)} = P(e_i \mid S_i)$, and because $e_1, \ldots, e_n$ are independent given $W$ we have (see Figure 3)

$$\begin{aligned} P(e_W \mid W) &= P(e_1 \mid W) \ldots P(e_n \mid W) \\ &= P(e_1 \mid S_1) \ldots P(e_n \mid S_n) \end{aligned}$$

Therefore

$$\begin{aligned} T_V(S) &= \sum_{W \setminus S} P(W) \frac{T_W(S_1)}{P(S_1)} \cdots \frac{T_W(S_n)}{P(S_n)} \\ &= \sum_{W \setminus S} P(W) P(e_i \mid S_1) \ldots P(e_n \mid S_n) \\ &= \sum_{W \setminus S} P(W) P(e_W \mid W) \\ &= \sum_{W \setminus S} P(W, e_W) \\ &= P(S, e_W) \end{aligned}$$

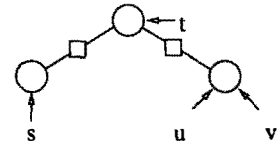

Figure 5: A set of findings, where $P(s, u, v), P(s, t, u)$ and $P(s, t, v)$ cannot be accessed directly.

## 4 Cautious entering of evidence

Consider the situation in Figure 5. We can access the complement of $s$. However, with the present techniques we cannot access the complements of $\{t\}, \{u\}$, and $\{v\}$. Fortunately there is an easy way out: Add dummy variables such that findings are always inserted to a leaf in the junction tree, and such that at most one finding is inserted in any node. The construction is illustrated in Figure 6.

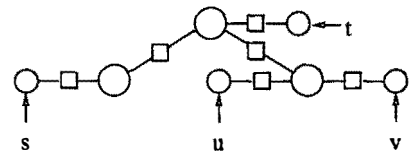

Figure 6: Now the probability of all complements of single findings can be accessed by cautious propagation.

This structural way of accessing complements of single findings through propagation can be achieved simpler by the following modification of the way that findings are inserted: Instead of changing the table in the clique when inserting the finding $f$, consider the finding as a message $F_f$, store it, and treat it in the same way as tables from neighbour separators. This way of handling findings corresponds to the one suggested by Dawid (1992).

Dawid used it in connection to the HUGIN propagation method to get $P(x \mid e \setminus \{x\})$. However, 0's can mess it up and in order for this to work always you have to use cautious propagation.

In the following we include cautious entering of evidence in the term *cautious propagation*.

## 5  $P(e')$s provided by cautious propagation

Let $S$ be any separator. $S$ divides the evidence $e$ into two sets $e_r$ and $e_l$, and cautious propagation provides both $P(e_r)$ and $P(e_l)$. Hence we for the situation in Figure 1 have access to the probability of the following

sets:

$\{s\}, e \setminus \{s\}, \{t\}, e \setminus \{t\}, \{u\}, e \setminus \{u\}, \{x\}, \{y\}, \{r\}, \{v\},$
$e \setminus \{v\}, \{w\}, e \setminus \{w\}, \{x, y\}, e \setminus \{s, y\}, \{z\}, e \setminus \{z\}, \{s, t\},$
$e \setminus \{s, t\}, \{u, v\}, e \setminus \{u, v\}, \{r, stuv, w\}, \{x, y, z\}$

Let $V$ be a clique with findings $f_1, \ldots, f_m$ entered and with adjacent separators $S_1, \ldots, S_n$. Let $e_i$ denote the evidence entered to the subtree containing $S_i$ but not $V$. Take for example the set $e_1 \cup e_2 \cup \{f_1\}$. Then

$$P(V, e_1, e_2, f_1) = P(V) P(e_1 \mid S_1) P(e_2 \mid S_2) F_{f_1}$$

and all the factors in the product are available local to $V$. Therefore $P(e_1 \cup e_2 \cup \{f_1\})$ is easy to calculate.

In general, cautious propagation gives access to the probability of any union of the sets $\{f_1\}, \ldots, \{f_m\}$, $e_1, \ldots, e_n$.

In Figure 1 we therefore also get access to the probability of

$e \setminus \{x\}, e \setminus \{y\}, e \setminus \{r\}, \{s, t, r\}, \{u, v, w, x, y, z\},$
$\{s, t, u, v\}, \{r, w, x, y, z\}, \{s, t, w\}, \{r, u, v, x, y, z\},$
$\{s, t, x, y, z\}, \{r, u, v, w\}, \{r, s, t, u, v\}, \{w, x, y, z\},$
$\{r, s, t, w\}, \{u, v, x, y, z\}, \{r, s, t, x, y, z\}, \{u, v, w\},$
$\{s, t, u, v, w\}, \{r, x, y, z\}, \{s, t, u, v, x, y, z\}, \{r, w\},$
$\{s, t, w, x, y, z, \}, \{r, u, v\}$

## 6 Complexity of cautious propagation

We shall compare cautious propagation (with cautious entering of evidence) and HUGIN propagation, which is the fastest known updating method (Shachter, Andersen & Szolovits 1991).

Let the junction tree have $n$ cliques (and $n-1$ separators), and assume that all cliques in the junction tree shall receive evidence.

The *space requirements* for cautious propagation is two extra tables for each separator and a way to store evidence. This will never require more than two times the space as for HUGIN propagation (most often considerably less).

As described in Section 3 cautious propagation requires the junction tree to be consistent before evidence is entered. This can be achieved by a HUGIN propagation; however, it may also be done by cautious propagation with tables of 1's in the separators. In this case, cautious propagation corresponds exactly to the propagation method suggested by Shafer & Shenoy (1990). When the propagation has terminated, the tables $P(S)$ for separators and $P(V)$ for cliques can be calculated from the tables stored (we shall revert to this later).

The methods are composed of the table operations *marginalization, multiplication* and *division*. To get insight into the time complexity of the methods we ignore that the time for the operations varies with table sizes. However, multiplication of $n$ tables is counted as $n - 1$ multiplications.

There are the following phases in the calculation of new probabilities:

a) Entering of evidence.

b) Propagation.

c) Calculation of marginals.

d) Reinitialization: Prepare the junction tree for a new set $e'$ of evidence. That is, all tables shall be conditioned on the evidence $e$ just entered. This phase is redundant when the case is closed, and also when there is no need for distinguishing between the two sets $e$ and $e'$ of evidence.

Re a) HUGIN: $n$ multiplications.
Cautious: No operations.

Re b) For both methods there are two marginalizations and two divisions for each link. For the multiplications we have

HUGIN (including entering of evidence):
Two multiplications for each clique except the root, where only one multiplication is necessary, $2n - 1$.

Cautious:
Let $V$ be a clique with $k$ neighbours. When a message has to be sent along a link, the table for $V$ shall be multiplied with $k - 1$ tables. In order to take care of all messages sent from $V$ we need $k(k-1)$ multiplications. For the entire propagation we therefore need

$$\text{Mult}_c = \sum_V \text{neighb}(V)(\text{neighb}(V) - 1)$$

propagations.

To analyse what that means, assume that $m$ cliques in the junction tree have $k$ neighbours and the remaining $n - m$ cliques are leaves. Since the graph is a tree we have $2(n-1) = km + (n-m)$ and we get $m = \frac{n-2}{k-1}$. This gives $\text{Mult}_C = k(n-2)$.

So, in the worst case $k = n - 1$ and $\text{Mult}_C = (n-1)(n-2)$; in the best case we have $\text{Mult}_C = 2(n-2)$.

For the rest of this analysis we assume that $k = 6$ (few junction trees are more heavily branched).

The operations above are performed on clique tables where findings have been entered. To establish this, the findings are multiplied on auxiliary tables. This expands the space requirement, so that it altogether is twice the requirement for HUGIN propagation.

Re c) The number of propagations are the same. We assume the number of variables to be $6n$, which gives $6n$ marginalizations.

Furthermore cautious propagation requires two multiplications for each link.

Re d) There is really nothing to do here in HUGIN propagation. What should be done is the normalization of all tables by dividing by $P(e)$, however, this can be postponed till later when table operations are performed. In cautious propagation the clique tables are calculated under c). For the separator tables we have (see Figure 4)

$$P(e, S) = P(e_V, e_W, S) = P(e_V \mid S) P(e_W \mid S) P(S)$$

So the three tables stored are multiplied. This requires $2(n-1)$ multiplications. Normalization is treated in the same way as for HUGIN propagation.

Counting koefficients of $n$ the complexity of HUGIN propagation amounts to $13n$, and under the assumptions made the complexity of cautious propagation is $21n$. So, altogether cautious propagation will very seldom take more than twice the time of HUGIN propagation.

## 7  An application: Sensitivity analysis

Sensitivity analysis is part of *explanation*, which has to do with explaining to a user how the system has arrived at its conclusions. Explanation for Bayesian networks has been studied systematically by Suermondt (1992), and Madigan & Mosurski (1993) have implemented various explanation facilities. Some of the questions to answer in connection to explanation concern the sensitivity of the conclusions to the particular evidence.

Let $h$ be a hypothesis (in the form of a particular configuration of states for some hypothesis variables), and let $e$ be the evidence entered. That is, $P(h \mid e)$ has been calculated, and now you would like to analyse the relation between the evidence and resulting probability.

**Definitions:** Let $e$ be evidence, $h$ a hypothesis, and let $\theta_1, \theta_2, \theta_3$ be predefined thresholds. Evidence $e' \subseteq e$ is

- *important* if $\frac{P(h \mid e \setminus e')}{P(h \mid e)} < 1 - \theta_1$

- *sufficient* if $\frac{P(h \mid e')}{P(h \mid e)} > 1 - \theta_2$

- *minimal sufficient* if $e'$ is sufficient, but no proper subset of $e'$ is so

- *crucial* if $e'$ is a subset of any sufficient set

- *decisive* if $P(h \mid e') > 1 - \theta_3$

As can be seen from the definitions above, the heart of sensitivity analysis is to calculate $P(h \mid e')$ for subsets $e'$ of $e$. If $e$ is not a very small set we cannot hope for an easy way of calculating $P(h \mid e')$ for all subsets. However, with cautious propagation (and cautious entering of evidence) the probability for a crucial number of subsets can be achieved.

Cautious propagation yields $P(e')$ for subsets as described in Sections 3 and 4. Now, enter $h$ as evidence to a clean junction tree, and HUGIN propagate such that all tables are conditioned on $h$ (this also yields $P(h)$). Next $e$ is entered cautiously and cautious propagation is performed. This gives access to $P(e' \mid h)$ for the same subsets as before, and Bayes' formula gives $P(h \mid e')$. (Jensen, Aldenryd & Jensen 1995) gives some examples of how cautious propagation can be used in sensitivity analysis.

Another natural set of questions in sensitivity analysis are *What-if-questions*: If the finding on the variable $X$ had been $y$ instead of $x$, what would $P(h \mid e \cup \{y\} \setminus \{x\})$ be?

This type of question can also easily be answered through cautious propagation and cautious evidence entering. Let namely $X$ be a member of $V$, and let a finding on $X$ be $x$. After cautious propagation the situation is, that local to $V$ you have $P(V), P(e_i \mid S_i)$ for all adjacent separators $S_i$, and tables $F_f$ for the findings $f$ to multiply on $P(V)$. It is then easy to substitute $F_x$ with $F_y$ and calculate $P(V, e \cup \{y\} \setminus \{x\})$ as the product of all tables local to $V$, and finally marginalize $V$ out to get $P(e \cup \{y\} \setminus \{x\})$. The same is done with the junction tree conditioned on $h$ to get $P(e \cup \{y\} \setminus \{x\} \mid h)$, and Bayes' formula yields $P(h \mid e \cup \{y\} \setminus \{x\})$.

Note that the analysis above did not require any extra propagations.

### Acknowledgements

Thanks to the ODIN group at Aalborg University (http:www.iesd.auc.dk/odin), in particular to Søren Dittmer for valuable discussions. Thanks also to Søren Aldenryd and Klaus B. Jensen for their contributions to the part on sensitivity analysis.